\documentclass[10pt,twocolumn,letterpaper]{article}

\usepackage[pagenumbers]{configs/cvpr} 

\usepackage{color}
\usepackage{amsthm}
\usepackage{wrapfig}
\usepackage{array}
\usepackage{newlfont} %
\usepackage{textcomp}
\usepackage{multirow}
\usepackage{hhline}
\usepackage{tabularx}
\usepackage{bigstrut}
\usepackage{afterpage}
\usepackage{algorithmic}
\usepackage{microtype}
\usepackage{mathtools}
\usepackage{booktabs} %
\usepackage[ruled]{algorithm2e} %
\usepackage{dingbat}
\usepackage[accsupp]{axessibility}
\PassOptionsToPackage{bookmarks=false}{hyperref}

\usepackage{comment}
\usepackage{times}
\usepackage{epsfig}
\usepackage{graphicx}
\usepackage{amsmath}
\usepackage{amssymb}
\usepackage{cuted}
\usepackage{capt-of}
\usepackage{float}
\usepackage{enumitem}
\usepackage{balance}
\usepackage{bbding} 
\usepackage[normalem]{ulem} 

\usepackage[dvipsnames]{xcolor}
\definecolor{mypink3}{cmyk}{0, 0.7808, 0.4429, 0.1412}
\definecolor{mygray}{gray}{0.6}

\usepackage[toc,page]{appendix}
\usepackage[numbers,sort,compress]{natbib}

\usepackage{gensymb}

\usepackage[pagebackref,breaklinks,colorlinks]{hyperref}

\usepackage{color, colortbl}

\usepackage{amssymb}
\usepackage{pifont}

\usepackage[tone, extra, safe]{tipa}
\usepackage{tipx}

\newcommand{\Fref}[1]{Fig.~\ref{#1}}

\newcommand{\tref}[1]{Tab.~\ref{#1}}

\newcommand{\fref}[1]{Fig.~\ref{#1}}

\newcommand{\ourtitle}{Generating Holistic 3D Human Motion from Speech}

\newcommand{\bodymethodname}{\textcolor{black}{SHOW}\xspace}
\newcommand{\bodyfullmethodname}{\textcolor{black}{``Synchronous Holistic Optimization in the Wild''}\xspace}
\newcommand{\speechmodelname}{\textcolor{black}{TalkSHOW}\xspace}

\definecolor{DeltaColor}{rgb}{0.039,0.73,0.71}
\definecolor{SigmaColor}{rgb}{0.98,0.45,0.0}
\definecolor{AlphaColor}{rgb}{0,0,0.8}
\definecolor{BetaColor}{rgb}{0.8,0,0.8}
\definecolor{GammaColor}{rgb}{0.514,0.34,0.224}
\definecolor{EpsilonColor}{rgb}{0.353,0.725,0.906}
\definecolor{PurpleColor}{rgb}{0.5,0,0.7}
\definecolor{OrangeColor}{rgb}{0.914,0.541,0.0.141}
\definecolor{GreenColor}{rgb}{0.137,0.573,0.565}
\definecolor{RedColor}{rgb}{0.949,0.275, 0.224}
\definecolor{LightCyan}{rgb}{0.88,1,1}
\definecolor{Gray}{gray}{0.3}
\definecolor{Strawberry}{rgb}{1,0.26,0.64}

\definecolor{BetaColor}{rgb}{0.8,0,0.8}

\definecolor{LightCyan}{rgb}{0.88,1,1}
\definecolor{lightgray}{rgb}{0.9,0.9,0.9}

\newcommand{\qheading}[1]{\noindent\textbf{#1}}
\newcommand{\tabincell}[2]{\begin{tabular}{@{}#1@{}}#2\end{tabular}} 

\definecolor{GreenColor}{rgb}{0.137,0.573,0.565}

\newcommand{\colorRef}[1]{\textcolor{black}{#1}} 
\usepackage[capitalize]{cleveref}
\crefname{figure}{\colorRef{Fig.}}{\colorRef{Figs.}}
\Crefname{figure}{\colorRef{Fig.}}{\colorRef{Figs.}}

\crefname{section}{\colorRef{Sec.}}{\colorRef{Secs.}}
\Crefname{section}{\colorRef{Sec.}}{\colorRef{Secs.}}
\Crefname{table}{\colorRef{Tab.}}{\colorRef{Tabs.}}
\crefname{table}{\colorRef{Tab.}}{\colorRef{Tabs.}}

\newcommand{\projectURL}{\url{https://talkshow.is.tue.mpg.de/}}

\renewcommand{\paragraph}[1]{\medskip\noindent\textbf{#1}}

\newcommand{\oursVideoLength}{27}

\def\cvprPaperID{961}
\def\confName{CVPR}
\def\confYear{2023}
\begin{document}

\title{\ourtitle}

\author{Hongwei Yi$^{1*}$\quad Hualin Liang$^{2*}$\quad Yifei Liu$^{2*}$\quad Qiong Cao$^{3\dagger}$ \\
 Yandong Wen$^{1}$\quad Timo Bolkart$^{1}$\quad Dacheng Tao$^{3}$\quad Michael J. Black$^{1\dagger}$ \\
$^{1}$Max Planck Institute for Intelligent Systems, T\"ubingen, Germany \\
$^2$South China University of Technology ~~~ $^3$JD Explore Academy \\
{\tt\small \{hongwei.yi, yandong.wen, timo.bolkart, black\}@tuebingen.mpg.de} \\
{\tt\small \{hualinliang3, yifei9697, mathqiong2012, dacheng.tao\}@gmail.com} 
}


\newcommand{\teaserCaption}{
\textbf{Speech-to-motion translation example.} Given a speech signal as input, our approach generates realistic, coherent, and diverse holistic body motions; that is, the body motion together with facial expressions and hand gestures. From top to bottom: the input audio, the corresponding transcript, video frames, and the generated motions. Note that the audio is the only input to our approach, while the transcript and video frames are just shown for reference. 

}

\twocolumn[{
    \renewcommand\twocolumn[1][]{#1}
    \vspace{-0.5cm}
    \maketitle
    \centering
    \begin{minipage}{1.00\textwidth}
        \centering
        \vspace{-0.8cm}
        \includegraphics[width=\textwidth]{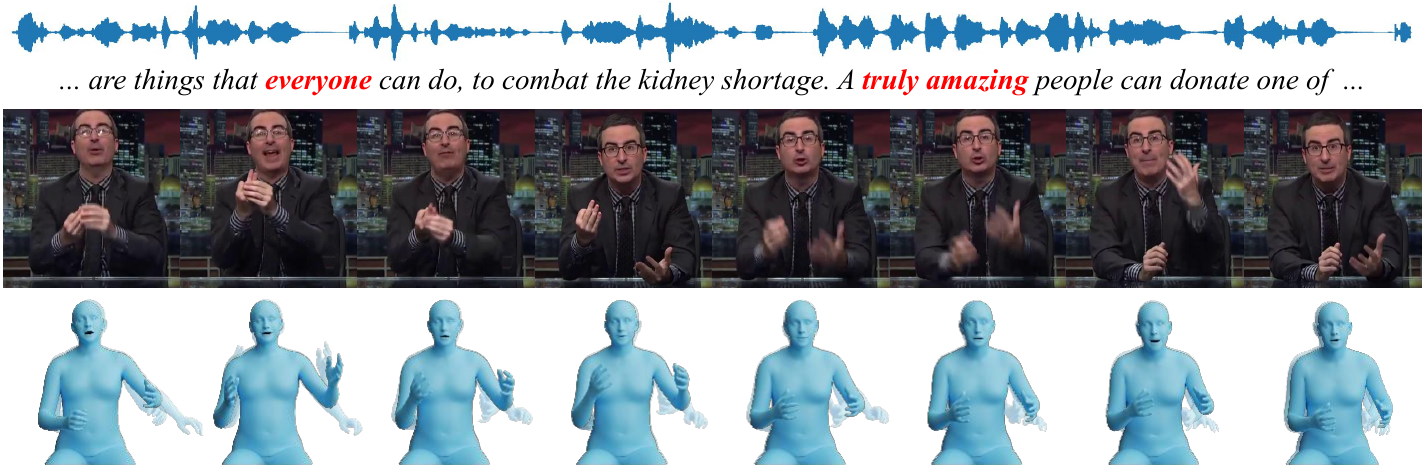}
    \end{minipage}
    \captionof{figure}{\teaserCaption}
    \vspace{0.5cm}
    \label{fig:teaser}
}]

\def\thefootnote{*}\footnotetext{Equal Contribution.}
\def\thefootnote{$\dagger$}\footnotetext{Joint Corresponding Authors.}

    


\begin{abstract}
This work addresses the problem of generating 3D holistic body motions from human speech. Given a speech recording, we synthesize sequences of 3D body poses, hand gestures, and facial expressions that are realistic and diverse. To achieve this, we first build a high-quality dataset of 3D holistic body meshes with synchronous speech. We then define a novel speech-to-motion generation framework in which the face, body, and hands are modeled separately. The separated modeling stems from the fact that face articulation strongly correlates with human speech, while body poses and hand gestures are less correlated. Specifically, we employ an autoencoder for face motions, and a compositional vector-quantized variational autoencoder (VQ-VAE) for the body and hand motions. The compositional VQ-VAE is key to generating diverse results. Additionally, we propose a cross-conditional autoregressive model that generates body poses and hand gestures, leading to coherent and realistic motions. Extensive experiments and user studies demonstrate that our proposed approach achieves state-of-the-art performance both qualitatively and quantitatively.
Our dataset and code are released for research purposes at \projectURL. 


\end{abstract}

\section{Introduction}
From linguistics and psychology we know that humans use body language to convey emotion and use gestures in communication \cite{goldin1999role,kendon2004gesture}.
Motion cues such as facial expression, body posture and hand movement all play a role. 
For instance, people may change their gestures when shifting to a new topic \citep{wagner2014gesture}, or wave their hands when greeting an audience. 
Recent methods have shown rapid progress on modeling the translation from human speech to body motion, and can be roughly divided into rule-based \citep{marsella2013virtual} and learning-based \citep{levine2009real,levine2010gesture,ferstl2018investigating,ginosar2019gestures,habibie2021learning, xu2022freeform} methods. Typically, the body motion in these methods is represented as the motion of a
3D mesh of the face/upper-body  \citep{karras2017audio, sadoughi2017meaningful, fan2022faceformer, richard2021meshtalk, ao2022rhythmic}, or 2D/3D landmarks of the face with  2D/3D joints of the hands and body \citep{ginosar2019gestures,habibie2021learning,xu2022freeform}. However, this is not sufficient to understand human behavior. Humans communicate with their bodies, hands and facial expressions  together. Capturing such coordinated activities as well as the full 3D surface in tune with speech is critical for virtual agents to behave realistically and interact with listeners meaningfully. 

In this work, we focus on generating the expressive 3D motion of person, including their body, hand gestures, and facial expressions, from speech alone; see \cref{fig:teaser}.
To do this, we must learn a cross-modal mapping between audio and 3D holistic body motion, which is very challenging in practice for several reasons. First, datasets of 3D holistic body meshes and synchronous speech recordings are scarce.
Acquiring them in the lab is expensive and doing so in the wild has not been possible.
Second, real humans often vary in shape, and their faces and hands are highly deformable. It is not trivial to generate both realistic and stable results of 3D holistic body meshes efficiently. 
Lastly, as different body parts correlate differently with speech signals, it is difficult to model the cross-modal mapping and generate realistic and diverse holistic body motions.

We address the above challenges and learn to model the conversational dynamics in a data-driven way. Firstly, to overcome the issue of data scarcity, we present a new set of 3D holistic body mesh annotations with synchronous audio from in-the-wild videos. This dataset was previously used for learning 2D/3D gesture modeling with 2D body keypoint annotations \citep{ginosar2019gestures} and 3D keypoint annotations of the holistic body \citep{habibie2021learning} by applying existing models separately. Apart from facilitating speech and motion modeling, our dataset can also support broad research topics like realistic digital human rendering. Then, to support our data-driven approach to modeling speech-to-motion translation, an accurate holistic body mesh is needed. 
Existing methods have focused on capturing either the body shape and pose isolated from the hands and face \citep{bogo2016keep,kanazawa2018end,yi2019mmface,li2023niki,Fang2023learning,yi2022mover,sun2023trace, zhang2021pymaf}, or the different parts together, which often produces  unrealistic or unstable results, especially when applied to video sequences \citep{pavlakos2019expressive, feng2021collaborative, pymafx2022}. To solve this, we present \bodymethodname, which stands for \bodyfullmethodname. Specifically, \bodymethodname adapts SMPLify-X \cite{pavlakos2019expressive} to the videos of talking persons, and further improves it  in terms of stability, accuracy, and efficiency through careful design choices. Figure \ref{fig:rec_results} shows example reconstruction results. 

Lastly, we investigate the translation from audio to 3D holistic body motion represented as a 3D mesh (\cref{fig:teaser}). We propose \speechmodelname, the first approach to autoregressively synthesize realistic and diverse 3D body motions, hand gestures and facial expression of a talking person from speech. Motivated by the fact that the face (i.e.~mouth region) is strongly correlated with the audio signal, while the body and hands are less correlated, or even uncorrelated, \speechmodelname designs separate motion generators for different parts and gives each part full play. 
For the face part, to model the highly correlated nature of phoneme-to-lip motion, we design a simple encoder-decoder based face generator that encodes rich phoneme information by incorporating the pretrained wav2vec 2.0 \citep{baevski2020wav2vec}. On the other hand, to predict the non-deterministic body and hand motions, we devise a novel VQ-VAE \citep{van2017neural} based framework to learn a compositional quantized space of motion, which efficiently captures a diverse range of motions. With the learned discrete representation, we further propose a novel autoregressive model to predict a multinomial distribution of future motion, cross-conditioned between existing motions. From this, a wide range of motion modes representing coherent poses can be sampled, leading to realistic looking motion generation.


We quantitatively evaluate the realism and diversity of our synthesized motion compared to ground truth and baseline methods and ablations. To further corroborate our qualitative results, we evaluate our approach through an extensive user study. Both quantitative and qualitative studies demonstrate the state-of-the-art quality of our speech-synthesized full expressive 3D character animations. 


    

\begin{table*}
    \centering
    \footnotesize
    
    \begin{tabular}{l|cccccccc}
    \hline
    \tabincell{l}{Dataset} & \tabincell{c}{Head} & \tabincell{c}{Hand} & \tabincell{c}{Body} & \tabincell{c}{Holistic Body \\ Connection} &    \tabincell{c}{In-the-wild} &
    \tabincell{c}{Length} & 
    \tabincell{c}{Annotations} \\ 
    \hline
    
    
    Multiface~\cite{wuu2022multiface} & 3D mesh & \ding{56} & \ding{56} & \ding{56} & \ding{56} & - & {multi-camera} \\
    BIWI~\cite{fanelli_3-d_2010} & 3D mesh & \ding{56} & \ding{56} & \ding{56} & \ding{56} & - &{3D-scanner} \\ 
    VOCASET~\cite{cudeiro_capture_2019} & 3D mesh & \ding{56} & \ding{56} & \ding{56} & \ding{56} & - & {4D-scan} \\ 
    Takeuchi~et.al~\cite{takeuchi2017creating} & \ding{56} & \ding{56} & 3D keypoint & \ding{56} & \ding{56} & 5h & {MoCap} \\
    Trinity \cite{ferstl2018investigating} & \ding{56} & \ding{56} & 3D keypoint & \ding{56} & \ding{56} & 4h &{MoCap} \\ 
    \hline
    Yoon~et.al~\cite{yoon2019robots, yoon2020speech} & \ding{56} & \ding{56} & 3D keypoint & \ding{56} & \ding{52} & 52h &{p-GT} \\
    Speech2Gesture~\cite{ginosar2019gestures} & \ding{56} & 2D keypoint & 2D keypoint & \ding{56} & \ding{52} & 144h & {p-GT}\\ 
    Habibie~et.al~\cite{habibie2021learning} & 3D mesh & 3D keypoint & 3D keypoint & \ding{56} & \ding{52} & 33h & {p-GT}\\
    \hline
    \textbf{Ours} & 3D mesh & 3D mesh & 3D mesh & \ding{52} & \ding{52} & {\oursVideoLength}h & {p-GT} \\ 
    \hline
    \end{tabular}
    \caption{Comparison of different speech-to-motion datasets.} 
    
    
    \label{differences_in_related_work}
\end{table*}



\section{Related work}
\label{related-work}
\subsection{Holistic Body Reconstruction} 

Recent work addresses the problem of 3D holistic body mesh recovery \citep{joo2018total, xu2020ghum, ExPose:2020, pavlakos2019expressive, pymafx2022}. 
SMPLify-X \citep{pavlakos2019expressive} fits the parametric and expressive  SMPL-X model \citep{pavlakos2019expressive} to 2D keypoints obtained by off-the-shelf detectors (e.g.~OpenPose ~\citep{cao2017realtime}). 
PIXIE \citep{feng2021collaborative} directly regresses SMPL-X parameters using moderators that estimate the confidence of part-specific features. 
These features are fused and fed to independent regressors. 
PyMAF-X~\citep{pymafx2022} improves the body and hand estimation with spatial alignment attention.
In this work, we adapt the optimization-based SMPLify-X to videos of talking persons, and improve the stability and accuracy with several good engineering practices in terms of initialization, data term design, and regularization.

\subsection{Speech-to-Motion Datasets}
The existing speech-to-motion datasets can be roughly categorized as in-house and in-the-wild. The annotations of in-house datasets \citep{takeuchi2017creating,ferstl2018investigating,fanelli_3-d_2010,cudeiro_capture_2019,wuu2022multiface} are accurate but are limited in scale since the multi-camera systems used for data capture are expensive and labor intensive. Moreover, these datasets only provide annotations of the head \citep{fanelli_3-d_2010,wuu2022multiface,cudeiro_capture_2019} or body \citep{takeuchi2017creating,ferstl2018investigating}, 
and thus do not support whole-body generation. 
To learn richer and more diverse speaking styles and emotions, \citep{yoon2019robots,yoon2020speech} propose to use in-the-wild videos. The annotations are pseudo ground truth (p-GT) given by advanced reconstruction approaches, \emph{e.g.} \citep{cao2017realtime}. However, these released datasets use either 2D keypoints or 3D keypoints with 3D head mesh to represent the body. This disconnected representation limits the possible applications of the generated talking motions. In contrast to the aforementioned work, our dataset, reconstructed by SHOW, consists of holistic body meshes and synchronized speech, covering a wide range of body poses, hand gestures, and facial expressions. More details can be found in Table \ref{differences_in_related_work}.

\subsection{Holistic Body Motion Generation from Speech}

Holistic body motion generation from speech consists of three body parts motion generation, i.e., faces, hands, and bodies.
Existing {3D} talking face generation methods \cite{zhou2018visemenet, cudeiro_capture_2019, richard2021meshtalk, fan2022faceformer} rely heavily on   4D face scan datasets  for training \cite{fanelli_3-d_2010, cudeiro_capture_2019, richard2021meshtalk}. 
There are many attempts to perform body motion generation, and these can be divided into rule-based and learning-based methods.  
Rule-based methods \citep{cassell2001beat, kopp2004synthesizing, levine2010gesture, poggi2005greta} map the input speech to pre-collected body motion ``units" with manually designed rules. They are explainable and controllable but it is expensive to create complex, realistic, motion patterns. 
%
Learning-based body motion generation approaches \citep{yoon2019robots, ginosar2019gestures, kucherenko2019analyzing, ahuja2020no, liao2020speech2video, bhattacharya2021speech, kucherenko2020gesticulator} have advanced significantly in part due to publicly released synchronous speech and body motion datasets \cite{takeuchi2017creating, yoon2019robots, yoon2020speech, ginosar2019gestures, habibie2021learning, li2021audio2gestures}. 
However, they only consider parts of the human body rather than the holistic body.
%
%
Most related to our work, 
Habibie et al. \citep{habibie2021learning} propose to generate 3D facial meshes and 3D keypoints of the body and hands from speech, but the generated faces, bodies and hands are disjoint. 
Also, these methods are deterministic, can not generate diverse motions when given the same speech recording.
There are a few attempts to incorporate the diversity into motion generation using GANs \citep{ahuja2020style, yoon2020speech, liu2022learning}, VAEs \citep{qian2021speech, xu2022freeform,li2021audio2gestures}, VQ-VAEs \citep{ao2022rhythmic, yazdian2022gesture2vec}, or normalising-flows \citep{alexanderson2020style}.
Nevertheless, the diversity of motions produced by these methods is inadequate. 

In contrast, TalkSHOW generates holistic body motions and models different body parts separately according to their natures: the face part is more correlated to the speech signal than body parts. 
TalkSHOW develops a simple deterministic encoder-decoder structure for mapping acoustic signals to facial expressions.
TalkSHOW adopts two VQ-VAEs to generate more diverse body and hand motions. 
This novel design allows the learned quantized space to be compositional and more expressive for conversational gestures. 
Compared with previous VQ-VAE-based methods \cite{siyao2022bailando, yazdian2022gesture2vec}, we design a cross-conditional autoregressive model to generate different body-part motions, which are more fluid and natural.

\section{Dataset} \label{dataset}
In this section, we introduce a high-quality audiovisual dataset, which consists of expressive 3D body meshes at 30fps, and their synchronized audio at a 22K sample rate. The 3D body meshes are reconstructed from in-the-wild monocular videos and are used as our pseudo ground truth (p-GT) in speech-to-motion generation. We provide detailed descriptions of this dataset in Sec.~\ref{data_descr} and highlight several good practices for obtaining more accurate p-GT from videos in Sec.~\ref{data_goodpractices}. Our experiments show that this dataset is  effective for training speech-to-motion models.

\subsection{Dataset Description} \label{data_descr}
The dataset is built from the in-the-wild talking videos of different people with various speaking styles. 
We use the same video sources from \cite{ginosar2019gestures} for straightforward comparisons with the previous work. To facilitate the subsequent 3D body reconstruction, we manually filter out  videos if they are in any following cases: (i) low resolution ($<$720p), (ii) occluded hand(s), or (iii) invalid download link. The filtering leads to a high-quality dataset of 26.9 hours from 4 speakers. For the mini-batch processing, the raw videos are cropped into short clips ($<$10 seconds). Direct comparisons to the existing datasets can be found in Table \ref{differences_in_related_work}.

Expressive 3D whole-body meshes are reconstructed from these videos and used as the p-GT. Specifically, the 3D holistic body meshes consist of face, hands, and bodies in a connected way, which is achieved by adopting a well-designed 3D topology from SMPL-X \cite{pavlakos2019expressive}. As a result, we represent the p-GT of the dataset as SMPL-X parameters. Given a video clip of $T$ frames, the p-GT comprises parameters of a shared body shape $\beta \in \mathbb{R}^{300}$, poses $\left\{\theta_{t}|\theta_{t}\in \mathbb{R}^{156}\right\}_{t=1}^{T}$, a shared camera pose $\theta^c\in \mathbb{R}^3$ and translation $\epsilon \in \mathbb{R}^3$, and facial expressions $\left\{\psi_{t}|\psi_{t} \in \mathbb{R}^{100}\right\}_{t=1}^{T}$. Here the pose $\theta_{t}$ includes the jaw pose $\theta^{jaw}_{t}\in \mathbb{R}^3$, the body pose $\theta^{b}_{t}\in \mathbb{R}^{63}$, and the hand pose $\theta^{h}_{t}\in \mathbb{R}^{90}$.

We note that this dataset can not only be used in speech and motion modeling, but also supports broad research topics like realistic digital human rendering and learning-based holistic body recovery from videos, etc. 

\newcommand{\reconstructedResultsCaption}{
The 3D holistic body reconstruction results from SMPLify-X, PIXIE, PyMAF-X, and ours. Compared to other methods, ours produces more accurate and stable results with details.}

\begin{figure}
    \centering
    \includegraphics[width=\columnwidth]{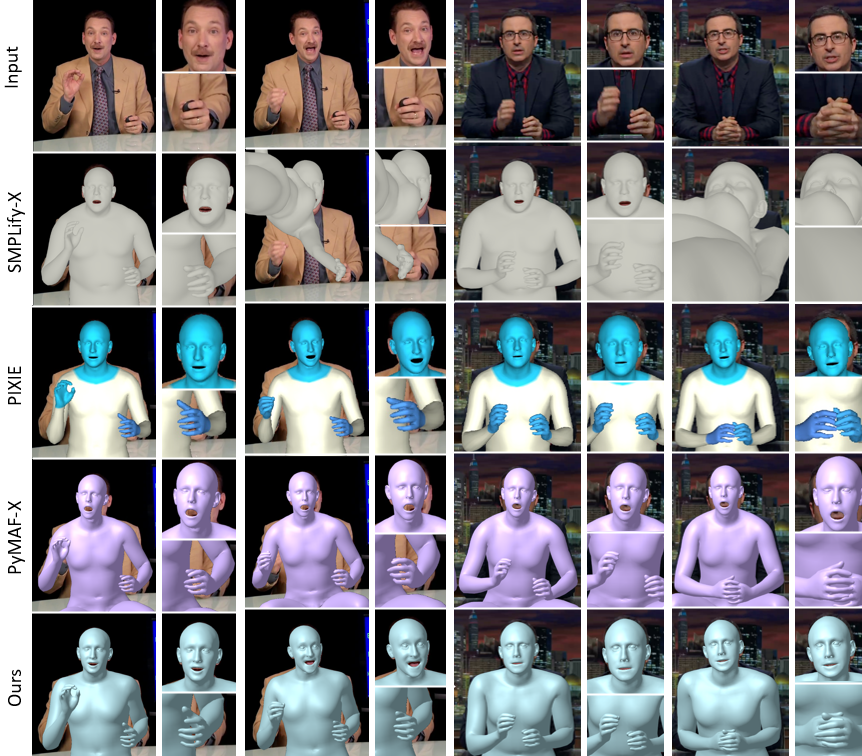}
    \caption{\reconstructedResultsCaption}
    \label{fig:rec_results}
\end{figure}

\subsection{Good Practices for Improving p-GT} \label{data_goodpractices}
%
In this section, we present \bodymethodname, which adapts SMPLify-X \cite{pavlakos2019expressive} to the videos of talking persons with several good practices, to improve the stability, accuracy, and efficiency in 3D whole-body reconstruction. 
In the following, we briefly summarize our efforts for improving the p-GT. See more details in the supplemental material.

\noindent \textbf{Initialization.} 
A good initialization can significantly accelerate and stabilize the SMPLify-X optimization. We apply several advanced regression-based approaches to the videos, and use the resulting predictions as the initial parameters of SMPLify-X. Specifically, PIXIE \cite{feng2021collaborative}, PyMAF-X \cite{pymafx2022}, and DECA \cite{feng2021learning} are used to initialize $\theta^b$, $\theta^h$, and $\theta^f$, respectively. The camera is assumed to be static, and its parameters $\theta^c$ and $\epsilon$ are estimated by PIXIE~\cite{feng2021collaborative} as well.

\noindent \textbf{Data Term.} 
The joint re-projection loss is the most important data objective function in SMPLify-X, as it optimizes the difference between joints extracted from the SMPL-X model, projected into the image, with joints predicted with OpenPose \cite{cao2017realtime}.
Here we extend the data term by incorporating body silhouettes from DeepLab V3, facial landmarks from MediaPipe \cite{kartynnik2019real}, and facial shapes from MICA \cite{zielonka2022mica}. 
Further, we use a photometric loss between the rendered faces and the input image to better capture facial details. 

\noindent \textbf{Regularization.} 
Different regularization terms in SMPLify-X prevent the reconstruction of unrealistic bodies. 
To derive more reasonable regularizations, we explicitly take information about the video  into account and make the following assumptions. First, the speaker in each video clip remains the same. This is further verified by a face recognition pipeline using the ArcFace model \cite{deng2019arcface}. So we can use consistent shape parameters $\beta$ to represent the holistic body shape. Second, the holistic body pose, facial expression, and environmental lighting in video clips change smoothly over time. This temporal smoothness assumption has proven useful in many previous approaches \cite{zielonka2022mica, yi2022mover}, and we observe similar improvements in our experiments. Third, the person's surface does  self-penetrate, which should be self-evident in the real world.

Overall, as shown in Figure \ref{fig:rec_results}, the p-GT can be significantly improved by incorporating the aforementioned practices. See more results in the supplemental video.

\section{Method}
\definecolor{newred}{RGB}{255, 0, 0}
\definecolor{newblue}{RGB}{65, 113, 156}
\definecolor{newgreen}{RGB}{169, 209, 142}
\newcommand{\generationResultsCaption}{Overview of the proposed \speechmodelname. We employ a simple encoder-decoder model for face motions, and a novel framework for body and hand motions. Specifically, this framework first learns VQ-VAEs on each piece separately to obtain a compositional quantized space. Then, we autoregressively predict the code indices of the body or hand motion following the \textcolor{red}{red} arrows orders. Our predictor is designed to be cross-conditioned between the body and hand motions to keep the synchronization of the holistic body. Lastly, we look up codes in the codebook according to indices and decode them to obtain the body and hand motion. 
Colors are utilized to differentiate codes within the same codebook, while various shades of these colors are used to separate body and hand codes.
Best viewed in color.
}
\begin{figure*}
    \centering
    \includegraphics[width=0.95\textwidth]{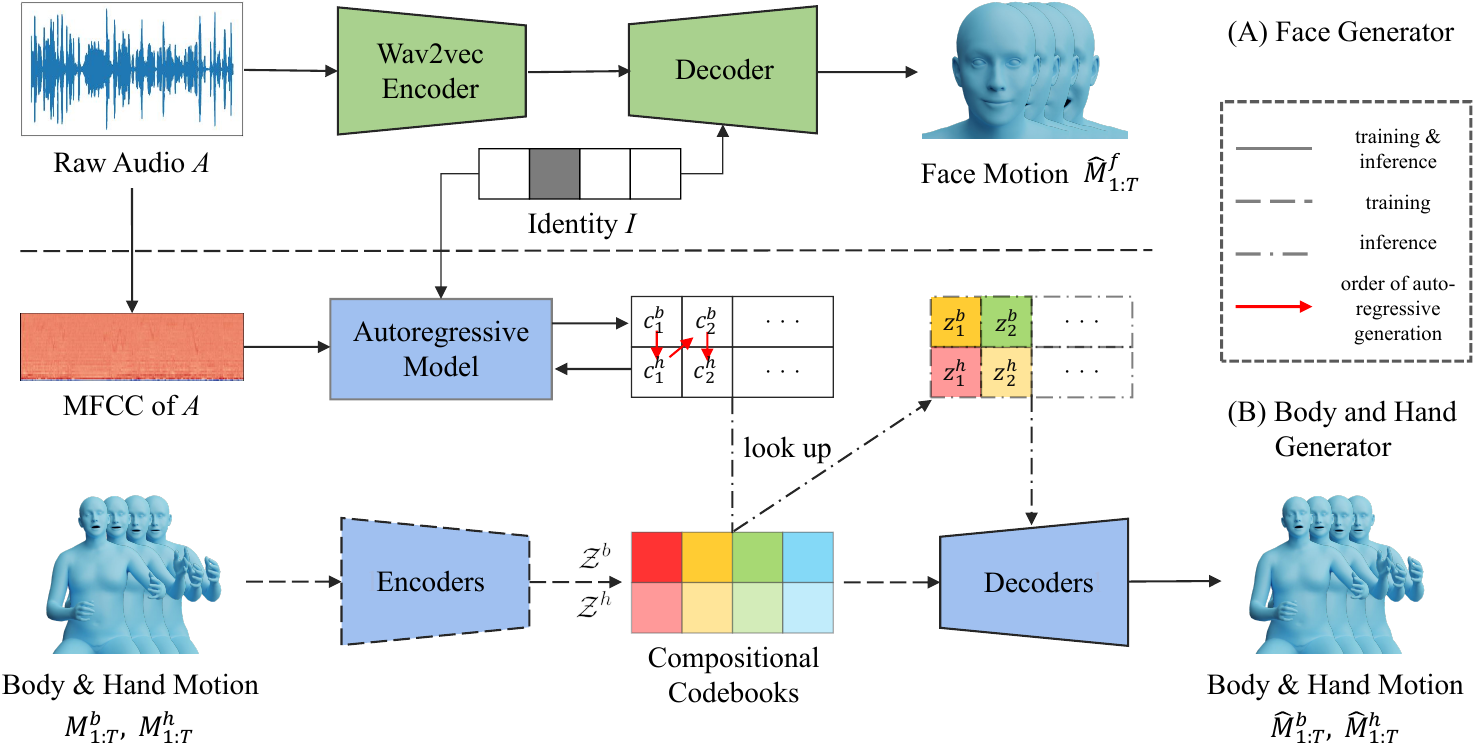}
    
    \caption{\generationResultsCaption}
    \vspace{-0.04in}
    \label{fig:architecture}
\end{figure*}

Given a speech recording, our goal is to generate conversational body poses, hand gestures as well as facial expressions that match the speech in a plausible way. Motivated by the fact that the face motion is highly correlated to the speech signal, while the body and hand parts are less correlated, we propose \speechmodelname, a novel framework that can model speech and different human parts separately. In the following, we present an encoder-decoder based face generator in Sec.~\ref{face-generator}, and a body and hand generator in Sec.~\ref{body-hands-generator}.





\subsection{Preliminary}
Let $M_{1:T}=\{m_t\}^T_{t=1}$ be a p-GT holistic motion (i.e., a temporal sequence of the holistic body poses $m_t=\{\theta_{t}, \psi_{t}\}$) provided in Sec.~\ref{dataset}.
We denote the motion of the face, body and hands as $M^f_{1:T}$, $M^b_{1:T}$ and $M^h_{1:T}$ respectively.
In particular, a facial motion $M^f_{1:T}$ is represented as a sequence of jaw poses and facial expression parameters $\{\theta_{t}^{jaw}, \psi_{t}\}^T_{t=1}$. And the body motion $M^b_{1:T}$ and the hands motion $M^h_{1:T}$ are denoted as a sequence of body poses $\{\theta_{t}^{b}\}^T_{t=1}$ and hand poses $\{\theta_{t}^{h}\}^T_{t=1}$, respectively.



\subsection{Face Generator} \label{face-generator}
Given a raw audio signal $A_{1:T}$ and speaker identity $I$, our face generator $G_F$ aims to generate an expressive facial motion $\widehat{M}^f_{1:T} = (\widehat{m}^f_1, \ldots, \widehat{m}^f_T) \in \mathbb{R}^{103 \times T}$ close to $M^f_{1:T} \in \mathbb{R}^{103 \times T}$. 


Figure \ref{fig:architecture} (A) illustrates our idea. In order to produce synchronized mouth motions \cite{fan2022faceformer}, we leverage a pretrained speech model, wav2vec 2.0 \citep{baevski2020wav2vec}. Specifically, the encoder consists of an audio feature extractor and a transformer encoder \citep{vaswani2017attention}, leading to a 768-dimensional speech representation. A linear projection layer is added on top of the encoder to reduce the feature dimension to 256. We then concatenate the audio feature with the speaker identity and feed them to the decoder. The speaker identity is represented as a one-hot vector $I \in \{{0,1}\} ^ {N_I}$, where $N_I$ is the number of speakers. Our decoder comprises six layers of temporal convolutional networks (TCNs) followed by a fully-connected layer. We train the encoder and decoder with an Mean Square Error (MSE) loss.





\subsection{Body and Hand Generator} \label{body-hands-generator}
Given an audio input, we aim to generate a realistic and diverse motion for the body and hands, \emph{i.e.} $\widehat{M}^b_{1:T} = (\widehat{m}^b_1, \ldots, \widehat{m}^b_T) \in \mathbb{R}^{63 \times T}$ and $\widehat{M}^h_{1:T} = (\widehat{m}^h_1, \ldots, \widehat{m}^h_T) \in \mathbb{R}^{90 \times T}$, respectively. 
Figure \ref{fig:architecture} (B) illustrates our idea. Instead of learning a direct mapping from audio to motion, we leverage the recent advances of VQ-VAE \cite{van2017neural} to learn a multi-mode distribution space for the body and hand motions. 
Specifically, we first encode and quantize the body and hand motions into two finite codebooks, from which we can sample a wide range of plausible body and hand combinations. Then, we introduce a novel cross-conditional autoregressive model over the learned codebooks, which allows us to predict diverse body and hand motions. 
Our predictor is designed to be cross-conditioned between the body and hands to keep the synchronization of the holistic body. Lastly, we obtain the future body/hand motion by decoding codebook indices sampled from the distribution. 

\paragraph{Representation.} We use 64-dimensional MFCC features \cite{sahidullah2012design} as the audio representation for body and hand motion generation, i.e., $A_{1:T} = (a_1, \ldots, a_T) \in \mathbb{R}^{64 \times T}$. Since body and hand gestures are more correlated to the rhythm and beat instead of phonemes, low-dimensional MFCC features are sufficient to produce plausible gestures from audio. Besides, considering that speakers often present different motion styles, we also leverage the modality of speaker identity $I$ to differentiate those styles. 

\paragraph{Compositional Quantized Motion Codebooks.}
The vanilla VQ-VAE learns a discrete codebook $\mathcal{Z} = \{z_i\}_{i=1}^{|\mathcal{Z}|}$ consisting of multiple vectors $z_i \in \mathbb{R}^{d_z}$ to quantize the latent space of the input. 
To further expand the range that the learned codebook can represent, we divide the body and hand motion into compositional pieces, i.e., body and hands, and learn VQ-VAEs on each piece separately.
By doing this, the body and hand motions are encoded and quantized into two separate finite codebooks $\mathcal{Z}^b = \{z^b_i\}_{i=1}^{|\mathcal{Z}_b|}$ and $\mathcal{Z}^h = \{z^h_j\}_{j=1}^{|\mathcal{Z}_h|}$, where $z^b_i, z^h_i \in \mathbb{R}^{d_z}$ with lengths $|\mathcal{Z}^b|$ and $|\mathcal{Z}^h|$ respectively. This approach enables us to obtain $|\mathcal{Z}^b|\times|\mathcal{Z}^h|$ different body-hand pose code pairs $(z^b_i, z^h_j)$ and expand the range of motion diversity.
In this scheme, given the body and hand motions $M^b_{1:T} \in \mathbb{R}^{63 \times T}$ and $M^h_{1:T} \in \mathbb{R}^{90 \times T}$ as input, we first encode them into the feature sequence $E^b_{1:\tau}=(e^b_1, \ldots, e^b_{\tau})\in \mathbb{R}^{64 \times \tau}$ and $E^h_{1:\tau}=(e^h_1, \ldots, e^h_{\tau})\in \mathbb{R}^{64 \times \tau}$. Here, $\tau=\frac{T}{w}$, where $w$ is the temporal window size, and $w$ consecutive poses correspond to one feature embedding. In this paper, we set $w=4$ to achieve a balance between inference speed and quality.
Then, we quantize the embedding by mapping it into the nearest code in the corresponding codebook:
\begin{equation}
\small
\begin{aligned}
    z^b_t &= \mathop{\arg\min_{z^b_k\in \mathcal{Z}^b}}\| e^b_t - z^b_k\| \in \mathbb{R}^{64}, \\ 
    z^h_t &= \mathop{\arg\min_{z^h_k\in \mathcal{Z}^h}}\| e^h_t - z^h_k\| \in \mathbb{R}^{64}.
\end{aligned}
\end{equation}
Finally, the quantized features $Z^b_{1:\tau} = (z^b_1, \dots, z^b_{\tau}) \in \mathbb{R}^{64\times \tau}$ and $Z^h_{1:\tau} = (z^h_1, \dots, z^h_{\tau}) \in \mathbb{R}^{64\times \tau}$ are fed into the decoder for the synthesis. 

We train the encoder, decoder, and codebook simultaneously with the following loss function:
\begin{equation}
\small
\begin{aligned}
    \mathcal{L}_{V Q}& =\mathcal{L}_{r e c}(M_{1:T}, \widehat{M}_{1:T})+\left\|\operatorname{sg}[E_{1:T}]-Z_{1:T}\right\| \\
    & +\beta\left\|E_{1:T}-\operatorname{sg}\left[Z_{1:T}\right]\right\|,
\end{aligned}
\end{equation}
where $\mathcal{L}_{r e c}$ is an MSE reconstruction loss, $\operatorname{sg}$ is a stop gradient operation \citep{chen2021exploring} that is used to calculate codebooks loss, and the third part is a ``commitment'' loss with a trade-off $\beta$.
\paragraph{Cross-Conditional Autoregressive Modeling.}
After we learn the compositional quantized codebooks, any body and hand motions can be represented as a sequence of codebook vectors via the encoder and quantization,
which is denoted as 
$C^b_{1:\tau} = (c^b_1, \dots, c^b_{\tau})\in \mathbb{R}^{|\mathcal{Z}^b| \times \tau}$ and $C^h_{1:\tau} = (c^h_1, \dots, c^h_{\tau})\in \mathbb{R}^{|\mathcal{Z}^h|\times \tau}$. 

Now, with the quantized motion representation, we design a temporal autoregressive model over it to predict the distribution of the possible next code, given the input audio embedding $A$ and existing motions. Besides, we enable the modality input of identity $I$ to distinguish different gesture styles. Because we model the body and hands independently, to keep the consistency of the holistic body and thus predict realistic gestures, we exploit the mutual information and design our model to be cross-conditioned between the body and hand motions. Specifically, following Bayes’ Rule, we model the joint probability of $C^b_{1:\tau}$ and $C^h_{1:\tau}$ as follows:
\begin{equation}
\small
\begin{aligned}
    p(C^b_{1:\tau}, C^h_{1:\tau} \mid A_{1:\tau}, I)=\prod_{t=1}^{\tau} 
    & p\left(c^b_t \mid c^b_{<t}, c^h_{<t}, a_{\leq t}, I\right) \\
    & p\left(c^h_t \mid c^b_{\leq t}, c^h_{<t}, a_{\leq t}, I\right).
    \end{aligned}
\end{equation}
Note that our cross-condition modeling between the body and hand motions makes the most of mutual information in two ways: (1) the current body/hand motion (i.e. $c^b_t$/$c^h_t$) depend on past body/hand motion information (i.e.~$c^b_{<t}$/$c^h_{<t}$); (2) we argue that the current body motion $c^b_{t}$ is also responsible for predicting the distribution of current hand motion. Such modeling guarantees the coherence of the body and hand motions as a whole and thus achieves realistic gestures. Gated PixelCNN \cite{van2016conditional} is adopted to model these quantities, in which the convolutional kernel is masked to make sure the model cannot read future information. During the training phase, the quantized body/hand motion representation concatenated with the audio and identity features is used for training. A teacher-forcing scheme and cross-entropy loss are adopted for the optimization. At inference, the model predicts multinomial distributions of the future body and hand motions, from which we can sample to acquire codebook indices for each motion. A codebook lookup is then conducted to retrieve the corresponding quantized element of motion, which we feed into the decoder for the final synthesis. Figure \ref{fig:architecture} (B) illustrates the pipeline. More training details are given in the supplemental material.

\section{Experiments}
    

\begin{table}
    \centering
    \small
    \resizebox{0.75\columnwidth}{!}{
    \begin{tabular}{l|cc}
    \hline
    \multirow{2}{*}{Method} 
    & \multicolumn{2}{c}{Face} \\
    & \tabincell{c}{L2 $\downarrow$} 
    & \tabincell{c}{LVD  $\downarrow$} \\ \hline 
    Habibie et al. \cite{habibie2021learning}    & 0.139    & 0.257      \\
    \textbf{TalkSHOW (Ours)}             & 0.130    & 0.248      \\
    \hline\hline
    \multirow{2}{*}{Method} 
    & \multicolumn{2}{c}{Body\&Hands} \\
    & \tabincell{c}{RS $\uparrow$} 
    & \tabincell{c}{Variation $\uparrow$} \\ \hline 
    Habibie et al. \cite{habibie2021learning}    & 0.146    & 0      \\
    Audio Encoder-Decoder           & 0.214    & 0  \\ 
    Audio VAE                      & 0.182    & 0.044 \\ 
    Audio+Motion VAE                & 0.240    & 0.176 \\ 
    \textbf{TalkSHOW (Ours)}             & 0.414    &0.821 \\ \hline
    \end{tabular}
    }
    
    \caption{Comparison to Habibie et al. \cite{habibie2021learning} and several baselines. $\uparrow$ indicates higher is better and $\downarrow$ indicates lower is better.}
    \label{tab:comparison_new}
\end{table}
We evaluate the ability of our method in generating body motions (i.e.~a sequence of poses) from the speech on the created dataset both quantitatively and qualitatively. Specifically, we choose video sequences longer than 3s and split them into 80\%/10\%/10\% for the train/val/test set. Several metrics are used to measure the realism and diversity of the generated motions including facial expression and poses. Furthermore, we conduct perceptual studies to assess the performance of our method. 


\subsection{Experimental Setup}

\paragraph{Evaluation Metrics.} Because we model face motion as a deterministic task and the body and hand motions as a non-deterministic task, we assess the generated motion in terms of the realism and the synchronization of face motion, and the realism and the diversity of body and hand motions. Specifically, the following metrics are adopted: 
\begin{itemize}
\item \textit{L2}: L2 distance between p-GT and generated facial landmarks, including jaw joints and lip shape \cite{zhou2020makelttalk, ng2022learning}.
\item \textit{LVD}: Landmark Velocity Difference calculates the velocity difference between p-GT and generated facial landmarks, which measures the synchronization between the input speech and the facial expression \cite{zhou2020makelttalk}. 
\item \textit{RS}: Score on the realism of the generated body and hand motions. Following \cite{xu2022freeform, aliakbarian2020stochastic}, we trained a binary classifier to discriminate real samples from fake ones and the prediction represents the realistic score. 
\item \textit{Variation}: As used in \cite{ng2022learning}, diversity is measured by the variance across 16 samples of body and hand motions.
\end{itemize}


\paragraph{Compared Methods.} We compare TalkSHOW to Habibie et al.~\cite{habibie2021learning}, a SOTA speech-to-motion method. Also, we compare several baselines for modeling body and hand motions when using the same face generator as ours:
\begin{itemize}
\item  \textbf{Audio Encoder-Decoder.} It encodes input audio and outputs motions;  this is used by \cite{ginosar2019gestures, habibie2021learning}.
\item \textbf{Audio VAE.} Given the input audio, the VAE-like structure encodes audio into a Gaussian distribution, and then the sampled audio is fed into the decoder, which transforms the sample into motions. 
\item \textbf{Audio+Motion VAE.} Given the input motion and audio, it adopts a VAE-like structure with two encoders to encode motion and audio into Gaussian distributions, respectively, and then the sampled motion and audio are concatenated and fed into the decoder for the synthesis. 
\end{itemize}

\subsection{Quantitative Analysis}
Table \ref{tab:comparison_new} shows the comparison results. We see that our method outperforms Habibie et al. \cite{habibie2021learning} across all metrics. Particularly, our method surpasses it in terms of L2 and LVD, which demonstrates the effectiveness of our face generator for generating realistic facial expressions. Also, our method significantly outperforms it in terms of variation, which demonstrates the powerful capacity to generate diverse body and hand motions resulting from our proposed compositional quantized motion representation. Moreover, regarding the realism (RS) for body and hand motions, we surpass Habibie et al. \cite{habibie2021learning} considerably, which confirms the effectiveness of our cross-conditioned autoregressive model in generating realistic motion. 

On the other hand, compared to VAE-based models, our method achieves large gains in both realism and diversity. 
In particular, we obtain much higher diversity. 
This indicates the advantage of the learned compositional quantized motion codebooks, which effectively memorize multiple motion modes of the body and hands and thus boost the diversity of the generated body and hand gestures.

\subsection{Qualitative Analysis}
\definecolor{newred}{RGB}{255, 0, 0}
\newcommand{\comparisonCaption}{
Our method generates diverse motions consistent with the rhythm of the input audio. For instance, we can generate different movements of hands corresponding to the strengthening tone of  ``\textit{\textcolor{newred}{But}}'' in the speech, e.g. using left hand only (top), right hand only (middle), or both hands (bottom).
}
\begin{figure}[t]
    \centering
    \includegraphics[width=0.8\columnwidth]{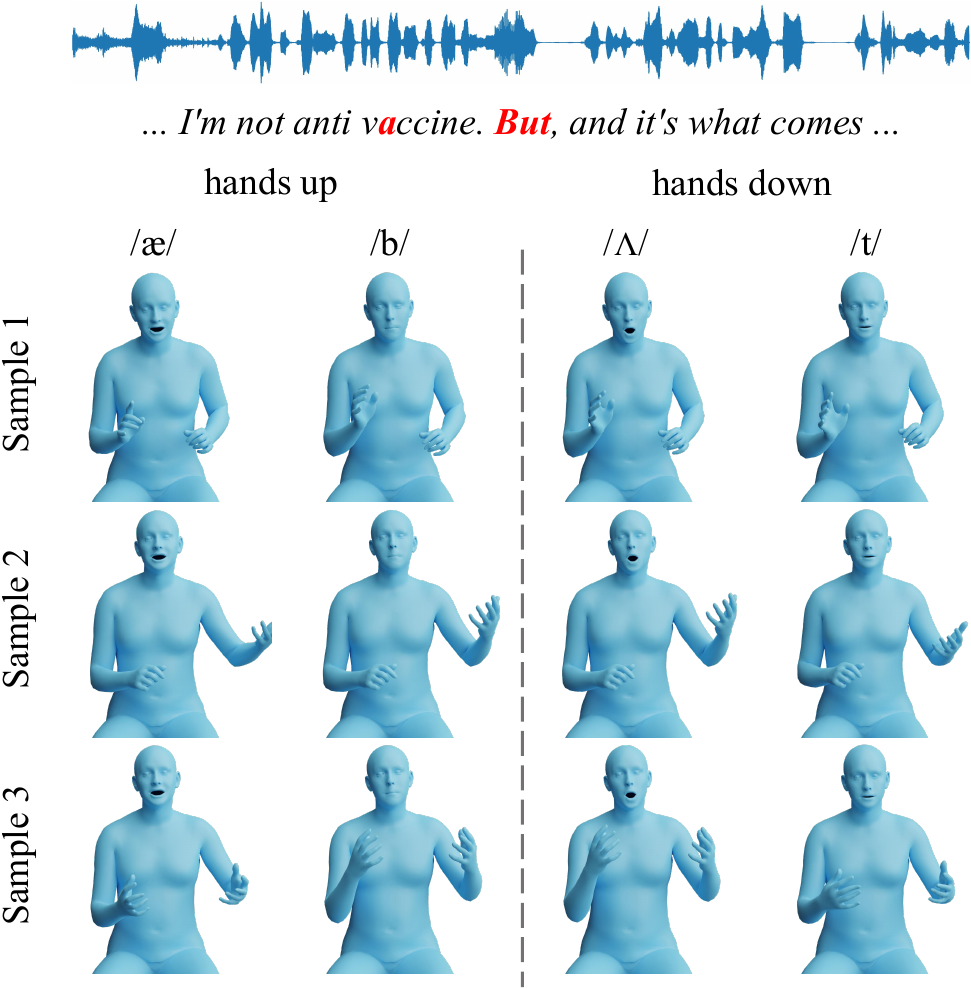}
    \vspace{-0.05in}
    \caption{\comparisonCaption}
    \label{fig:comparison}
\end{figure}

Figure \ref{fig:comparison} shows examples of our generated 3D holistic body motion from speech. 
We see that given the word ``\textit{\textcolor{red}{But}}'' from the speech which represents a strengthening tone of voice, our method generates plausible holistic body motions with hands up before saying ``\textit{\textcolor{red}{But}}'' and hands down after saying ``\textit{\textcolor{red}{But}}''.
Notably, the generated motions are diverse in many aspects, e.g. the range of motion and which hands to use, as shown in three different generated samples. 

\newcommand{\facedetailCaption}{
Given speech audio as input, our method generates facial expressions with accurate lip shapes.
}
\begin{figure}[t]
    \centering
    \includegraphics[width=0.8\columnwidth]{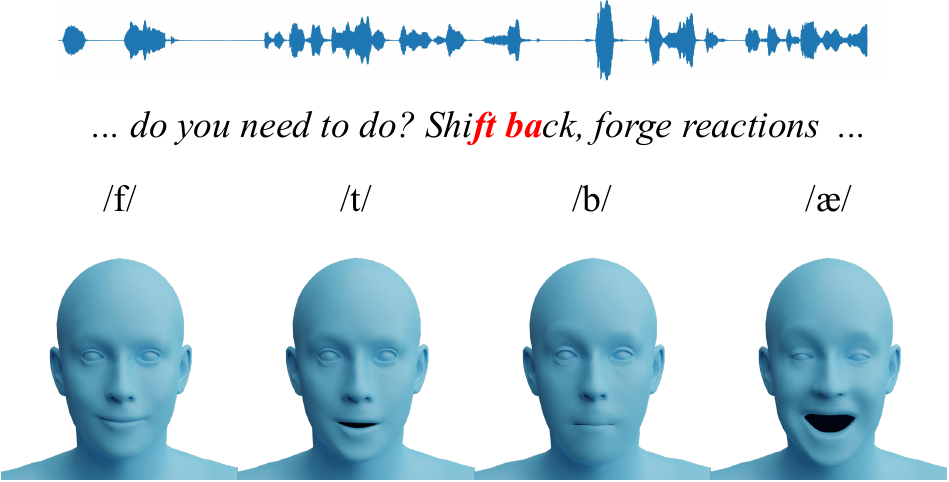}
    \vspace{-0.05in}
    \caption{\facedetailCaption}
    \label{fig:face_detail}
\end{figure}
\newcommand{\capacitycaption}{
The comparison of VQ-VAE and VQ-VAEs with compositional codebooks.}
\begin{figure}[t]
    \centering
    \includegraphics[width=0.26\textwidth]{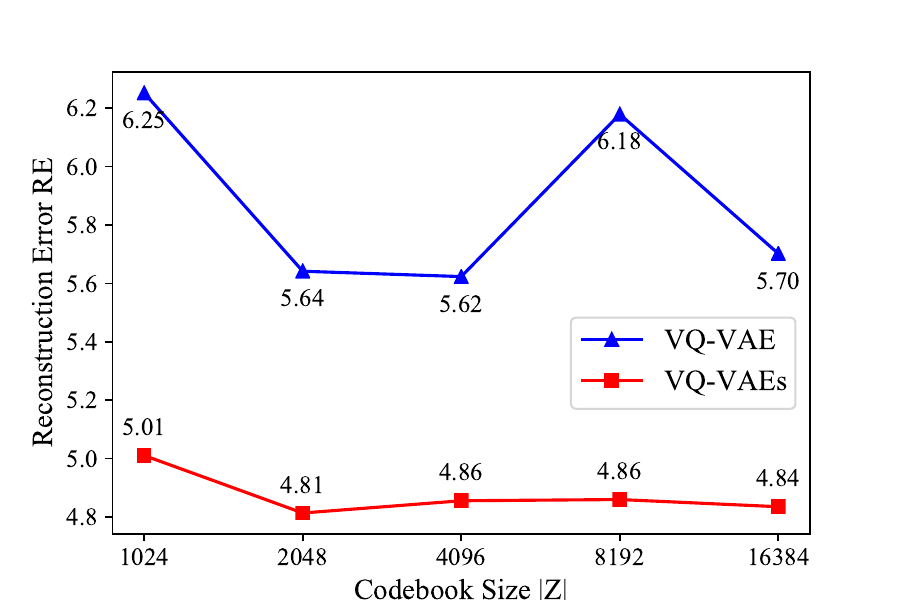}
    \caption{\capacitycaption}
    \label{fig:capacity}
\end{figure}
Figure \ref{fig:face_detail} illustrates the qualitative performance of our face generator. Our approach generates realistic face motions including consistent lip motions with the corresponding phonemes such as /f/, /t/, /b/, and /\ae/.
Furthermore, our method exhibits a robust generalization ability to unseen languages and various audio types, e.g. French and songs. 
Additional interesting examples can be found in our supplemental video.


\subsection{Model Ablation}
\paragraph{Effect of Wav2vec Feature in Face Generation.} We evaluate the effect of the wav2vec feature used in face generation compared to the MFCC feature. We add an extra encoder to increase the dimension of the MFCC feature from 64 to 256 for a fair comparison. The wav2vec-based model outperforms the MFCC-based model in both metrics (0.130 vs. 0.165 in L2 and 0.251 vs. 0.277 in LVD) due to its larger capacity for modeling the relationship between audio and phonemes. Moreover, we experimentally find that the wav2vec-based model can generalize well to unseen identities; see supplementary for more details.

\paragraph{Effect of Compositional Quantized Motion Codebooks.}
We analyze the capability of capturing the diverse motion modes represented in motion data by our proposed compositional quantized motion codebooks of VQ-VAEs.
To this end, we compare VQ-VAE with a single codebook. Reconstruction Error $RE$ is adopted as the metric, in which a lower reconstruction loss indicates a higher capacity. 
Figure \ref{fig:capacity} illustrates the results. We see that compared to VQ-VAE with a single codebook, VQ-VAEs with compositional codebooks yield consistently lower $RE$ across different codebook sizes. This demonstrates the effectiveness of the proposed compositional codebooks in modeling the diverse motion modes.



\paragraph{Effect of Cross-Conditional Modeling.} 
In contrast to cross-conditional modeling (w/ c-c), the model without cross condition (w/o c-c) generates body and hand motions independently.
Our method w/ c-c yields a higher realistic score than that w/o c-c ($0.414$ vs.~$0.409$), benefiting from the cross-conditional modeling between the body and hand motions which leads to generating more coherent and realistic motions. 
%
Our method w/ c-c attains a slight reduction in diversity ($0.821$ vs.~$0.922$ in variance), however the method w/o c-c leads to implausible body and hands combination.   

\subsection{Perceptual Study}
We conduct perceptual studies with Google Forms to evaluate our reconstruction and generation results, respectively. We randomly sample 40 videos in total with 10 videos from each speaker. Ten participants took part in the study. 

\paragraph{Reconstruction.}
We assess the quality of our holistic body reconstruct results against PyMAF-X~\cite{pymafx2022}, compared with the ground truth. Participants are asked to answer the following questions with Yes or No: Does the reconstructed face/hands/body/full-body match the input video? Table~\ref{tab:user_study_rec} reports the average percentage of answers that the reconstructed results match the input video. We see that our method outperforms PyMAF-X by a large margin.
\begin{table}
    \centering
    \footnotesize
    \resizebox{0.75\columnwidth}{!}{
    \begin{tabular}{l|cccc}
    \toprule
    \tabincell{l}{Method}       & \tabincell{c}{face} &
    \tabincell{c}{body} &
    \tabincell{c}{hands} &
    \tabincell{c}{holistic body}\\ 
    \hline
    PyMAF-X~\cite{pymafx2022} & 0.323 & 0.500 & 0.438 & 0.193 \\
    \hline
    \textbf{SHOW (ours)} & 0.898 & 0.738 & 0.800 & 0.768 \\
    \hline
    \end{tabular}
        }
        \vspace{-0.085in}
    \caption{
Perceptual study results on reconstruction. For each method, we report the average percentage of answers that the reconstructed results match the input video. 
    } 
    \label{tab:user_study_rec}
\end{table}

\paragraph{Holistic Body Motion Generation.}
We use A/B testing to evaluate our generation results, compared to the p-GT and Habibie et al.~\cite{habibie2021learning}. Specifically, participants are asked to answer the following questions with A or B: For the face/body\&hands/overall region, which one is a better match with the given speech? Table~\ref{tab:user_study_compare_with_gt} reports the average preference percentage of answers. We see that participants favor our method over Habibie et al. in terms of all the regions. Not surprisingly, participants perceive the p-GT better over both methods, with our method preferred by many more users.     

\begin{table}
    \centering
    \footnotesize
    \resizebox{0.9\columnwidth}{!}{
    \begin{tabular}{l|ccc}
    \toprule
    \tabincell{l}{Method}       
    & \tabincell{c}{face} &
    \tabincell{c}{body and hands} &
    \tabincell{c}{holistic body}\\ 
    \hline
     \cite{habibie2021learning} vs. p-GT   & 0.153 & 0.141 & 0.169 \\
    \textbf{TalkSHOW (Ours)} vs. p-GT  & 0.478 & 0.464 & 0.458 \\
    \hline
    \textbf{TalkSHOW (Ours)}  vs. \cite{habibie2021learning} & 0.888 & 0.910 & 0.913 \\
    \hline
    \end{tabular}}
    \vspace{-0.05in}
    \caption{
    Perceptual study of motion generation. We use A/B testing and report the percentage of answers where A is preferred over B.  
    }
    \label{tab:user_study_compare_with_gt}
\end{table}

\section{Conclusion}
In this work, we propose \speechmodelname, the first approach to generate 3D holistic body meshes from speech. We devise a simple and effective encoder-decoder for realistic face generation with accurate lip shape. For body and hands, we enable diverse generation and coherent prediction with compositional VQ-VAE and cross-conditional modeling, respectively. Moreover, we contribute a new set of accurate 3D holistic body meshes with synchronous audios from in-the-wild videos. The annotations are obtained by an empirical approach designed for videos. 
Experimental results demonstrate that our proposed approach achieves state-of-the-art performance both qualitatively and quantitatively. 





\medskip

\noindent
{\qheading{Acknowledgments.}
We thank W.~Zielonka and J.~Thies for helping us incorporate MICA into SHOW, C.~Ding, H.~Jiang, Y.~Feng, Z.~Liu, and W.~Liu for insightful discussions, and B.~Pellkofer for IT support.
This work was supported by the German Federal Ministry of Education and Research (BMBF): Tübingen AI Center, FKZ: 01IS18039B.
}

{\small \qheading{Disclosure.}
\href{https://files.is.tue.mpg.de/black/CoI_CVPR_2023.txt}{https://files.is.tue.mpg.de/black/CoI\_CVPR\_2023.txt}}


\bibliography{ref.bib}
\bibliographystyle{plainnat}

\clearpage

\newpage
\begin{appendices} \label{appendices}


\section{Dataset}

\subsection{Dataset Description}
Our dataset is built from the in-the-wild talking videos of four persons with various poses. The dataset contains high-quality 3D holistic body mesh annotations that are reconstructed from video clips of 26.9 hours in total. Each clip is less than 10 seconds. \fref{fig:datasets_distrbution} illustrates the distributions of video durations from different characters.

\newcommand{\distributeCaption}{
The distribution of the number of short clips for each character (0-10 seconds) of different speakers.
}

\begin{figure}[h] 
    \includegraphics[width=\columnwidth]{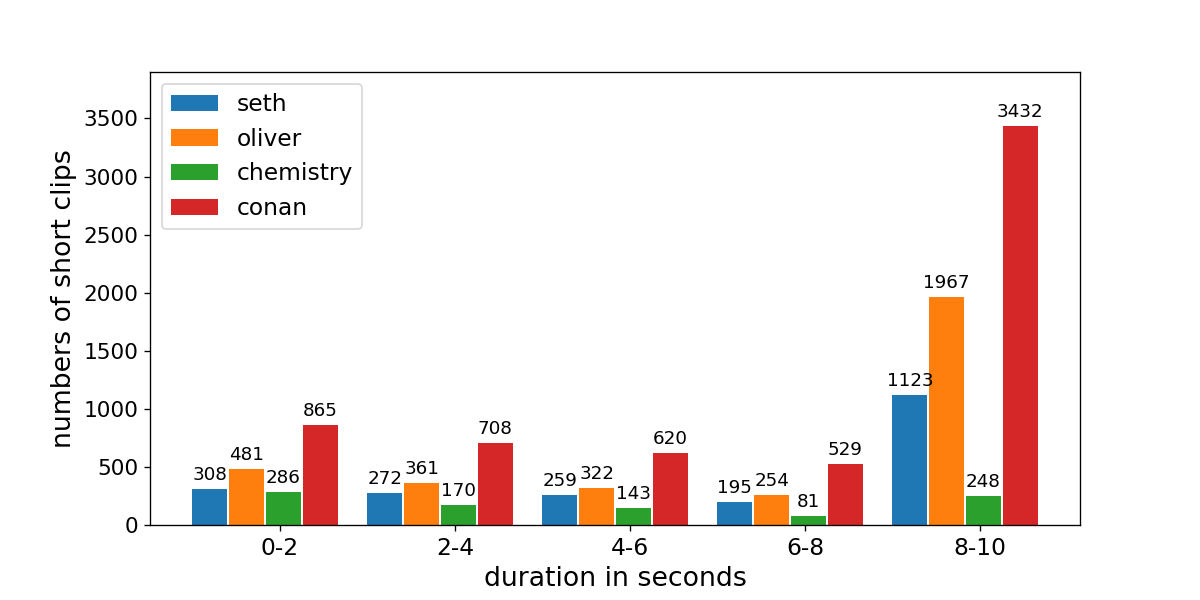}
    \caption{\distributeCaption}
    \label{fig:datasets_distrbution}
\end{figure}

\subsection{Good Practices for Improving p-GT}

\newcommand{\architectureRecCaption}{
The architecture of \bodymethodname. It consists of initialization and optimization modules. Specifically, given an input the image sequence, firstly, PIXIE~\cite{feng2021collaborative}, DECA~\cite{deng2019accurate} and PyMAF-X~\cite{zhang2021pymaf} are used to initialize the parameters of SMPL-X. Secondly, the optimization routine incorporates body silhouettes from DeepLab V3~\cite{DBLP:journals/corr/ChenPSA17}, facial landmarks from MediaPipe~\cite{kartynnik2019real}, and facial shapes from MICA~\cite{zielonka2022mica}. Then, it uses a photometric loss between the rendered faces and the input image to better capture facial details. Lastly, \bodymethodname outputs the final results.

}

\begin{figure*} 
    \label{arch_reconstruction}
    \centering
    \includegraphics[width=\textwidth]{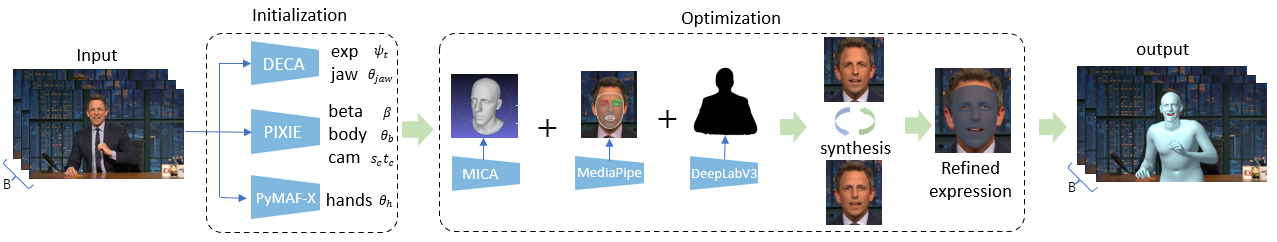}
    \caption{\architectureRecCaption}
    \label{fig:architecture_rec}
\end{figure*}

\paragraph{Preliminary.}
The 3D holistic body meshes consist of face, hands, and body, which is achieved by adopting SMPL-X \cite{pavlakos2019expressive}. It uses standard vertex-based linear blend skinning with learned corrective blend shapes and has $N=10475$ vertices and $K=67$ joints. Let $W$ be the linear blend skinning function, the predicted mesh vertices can be represented as $v = W(\theta,\psi,\beta) \in \mathbb{R}^{N \times 3}$. Let $V=\{v_t|v_t \in \mathbb{R}^{N \times 3}\}^T_{t=1}$ and $J=\{j_t|j_t \in \mathbb{R}^{67}\}^T_{t=1}$ be the temporal sequence of mesh vertices and its 3D joint locations regressed from a linear regressor. We also denote 
$P^{b}=\{p^{b}_t|p^{b}_t \in \mathbb{R}^{32}\}^T_{t=1}$ and $P^{h}=\{p^{h}_t|p^{h}_t \in \mathbb{R}^{24}\}^T_{t=1}$ as the temporal sequence of the coefficients of the latent space of VPoser and low dimensional pose space after principal component analysis (PCA) for the body and hands respectively. For time interval $[1:t]$, $V_{1:t} = (v_1,...,v_t)$, $J_{1:t} = (j_1,...,j_t)$, $P^{b}_{1:t} = (p^{b}_{1},...,p^{b}_{t})$ and $P^{h}_{1:t} = (p^{h}_{1},...,p^{h}_{t})$ represent segments of mesh vertices, 3D joints, body pose, and hand pose, respectively. Note that we use a fixed pose (sitting or standing) for the invisible lower body.
And in a temporal sequence of the p-GT holistic motions $m_i$, at each time step $t$, the facial representation $m^f_t=[\theta^{f}_{t}, \psi_t]\in \mathbb{R}^{103}$ is a concatenation of jaw orientation and expression, and the body and hand motions are respectively represented by their poses $m^b_t=\theta^{b}_t\in \mathbb{R}^{63}$ and $m^h_t=\theta^h_t\in \mathbb{R}^{90}$.


\paragraph{Initialization.}
Since optimization-based methods are often slow and sensitive to the initialization. In contrast, regression-based methods tend to give a reasonable, but not well pixel-aligned results. Therefore, we use the results from PIXIE~\cite{feng2021collaborative} and PyMAF-X~\cite{zhang2021pymaf} to initialize the parameters of body and hand pose, respectively. Results from DECA~\cite{feng2021learning} are used to initialize the  parameters of jaw pose and facial expression.


\paragraph{Data terms.}
We extend the data term by incorporating body silhouettes, facial landmarks, facial shapes, and facial details.

Firstly, to deal with the imperfect 2D landmarks by Openpose~\cite{cao2017realtime}, we introduce the silhouette constraint to encourage the rendered SMPL-X body to be inside the human body mask. Ground-truth person segmentations are expensive to obtain for in-the-wild datasets. Hence, we employ an off-the-shelf segmentation model, Deeplab V3~\cite{DBLP:journals/corr/ChenPSA17} to generate p-GT person semantics mask $M_{sil} \in \mathbb{R}^{T \times h \times w}$, where H and W are the height and width of the input image. Pytorch3D is used as the differential renderer to process the rendered pixels of all mesh triangles, leading to the predicted semantics mask $\widehat{M}_{sil} \in \mathbb{R}^{T \times h \times w}$. The silhouette loss term is given by:

\begin{equation}
\mathcal{L}_{sil} = \sum || d(\widehat{M}_{sil}) \odot d_{edt}(g(M_{sil})) ||_2,
\end{equation} 
where $g(x)=MaxPool(x)-x$ is a function for detecting the edge of the binary mask.  $d_{edt}$ is a distance function to calculate the smallest Euclidean distance from the background point to the silhouette boundary.


Secondly, to get a better facial geometry in SMPL-X, we minimize the difference between the facial shape in SMPL-X and the reconstructed facial shape from MICA~\cite{zielonka2022mica}. We term this as a facial shape objective $\mathcal{L}_{FS}$ given by:
\begin{equation}
\mathcal{L}_{FS} = || {M}_{g1}({V_{SMPL-X}}) - M_{g2}({V}_{MICA} + {t}_{FS}) ||_2,\end{equation}
where $V_{SMPL-X} \in \mathbb{R}^{N \times 3}$ is the SMPL-X vertices at neutral pose (i.e. $\theta = 0,\psi = 0$). ${V}_{MICA} \in \mathbb{R}^{5023 \times 3}$ is the MICA shape, and ${t}_{FS} \in \mathbb{R}^{3}$ is the offset of ${V}_{MICA}$ from $V_{SMPL-X}$. $M_{g1}$ and $M_{g2}$ are functions that maps the original mesh vertices of $V_{SMPL-X}$ and ${V}_{MICA}$ to the corresponding $1787$ vertices of frontal face part, respectively.





Thirdly, to get better facial expression, we use MediaPipe~\cite{kartynnik2019real} to extract $105$ of $468$ dense 2D facial landmarks for each image. The loss term $\mathcal{L}_{FE}$ is calculated as:
\begin{equation}
    \mathcal{L}_{FE} = \sum_{t}|| {U}_{1:t} - \widehat{U}_{1:t} ||_2,
\end{equation}
where ${U}_{1:t}$ and $\widehat{U}_{i}$ are temporal segments of landmarks from MediaPipe~\cite{kartynnik2019real} and the 2D projection of the corresponding 3D joints $J_{1:t}$, respectively.

Lastly, to obtain high-frequency resolution facial details, we employ face expression tracking to monocular RGB images in a self-supervised fashion. Specially, we follow ~\cite{feng2021learning,zielonka2022mica} to reconstruct the face jointly with an illumination model based on spherical harmonics and a Lambertian material assumption: 

\begin{equation}
\mathcal{L}_{FR} = \sum_{t}|| I_r(\mathcal{M}_{S2F}(V_{1:t})) - I_{1:t}^{head} ||_2,
\end{equation} 
where $\mathcal{M}_{S2F}$ is a function that selects the head part of ${V_{1:t}}$. $I_r$ is the forward pass of differential rendering. $I_{1:t}^{head}$ is the cropped head image from input image. Note that we choose different scales (e.g. $256,512,1024$) for different stages in the optimization procedure.  


\paragraph{Regularization.}
Different regularization terms in SMPLify-X prevent the reconstruction of unrealistic bodies. To derive more reasonable regularization terms, we explicitly take the video prior into account.

To reduce the jittery results caused by the noisy 2D detected keypoints, we introduce a smooth term for body and motion poses ($P^{b}$ and $P^{h}$). They are defined as:

\begin{equation}
    \mathcal{M}_{b} = \sum_{t}|| P^{b}_{2:t} - P^{b}_{1:t-1} ||_2,
\end{equation}
\begin{equation}
    \mathcal{M}_{h} = \sum_{t}|| P^{h}_{2:t} - P^{h}_{1:t-1} ||_2.
\end{equation}

We also add constant-velocity smooth term $\mathcal{M}_{j}$ on ${J}$:
\begin{equation}
    \mathcal{M}_{j} = \sum_{t}|| J_{3:t} + J_{1:t-3} - 2 \times J_{2:t-2} ||_2,
\end{equation}

Furthermore, to prevent the inter-penetration of two hands, we use Collision Penalizer~\cite{pavlakos2019expressive} and denote this loss term as $L_{pen}$.

\paragraph{Training Losses.} The final objective function is given by:
\begin{multline} 
E(\beta, \{\theta\}_{t=1}^{T}, \{\psi\}_{t=1}^{T}, \psi_{light}, \psi_{lbs}, t_{FS}) = \\ \sum_{t=1}^{T}(E_{SMPLify-X}(t)) + \lambda_{FE}\mathcal{L}_{FE} + 
\lambda_{FS}\mathcal{L}_{FS} + 
\lambda_{FR}\mathcal{L}_{FR} + \\
\lambda_{mb}\mathcal{M}_{b} +
\lambda_{mh}\mathcal{M}_{h} +
\lambda_{mj}\mathcal{M}_{j} +
\lambda_{sil}\mathcal{L}_{sil} 
+ \lambda_{pen}\mathcal{L}_{pen},
\end{multline} 
where $\psi_{light} \in R^{3}$ is the spherical harmonic coefficients representing the environmental illumination. $\psi_{lbs} \in R^{128}$ is the linear blend skinning parameters of albedo model. $E_{SMPLify-X}(t)$ is the basic prior on single image as describe in ~\cite{pavlakos2019expressive}. Weights $\lambda$ steer the influence of each term. 

\paragraph{Optimization.}
Following~\cite{pavlakos2019expressive}, we adopt the Limited-memory BFGS ~\cite{nocedal2006nonlinear} with strong wolfe line search for optimization. An iterative fitting routine is used for better fitting. With proper initialization, we minimize the objective function using a five-stage fitting procedure to avoid the local minima trap and reduce the optimization time. The learning rate is set to 1. As the required GPU memory increases dramatically with the image batch size for neural rendering, we use a mini-batch of $50$ on NVIDIA Tesla V100.



\newcommand{\reconstructedResultsCaptionv}{
The 3D holistic body reconstructions of four subjects from SMPLify-X, PIXIE, PyMAF-X, and ours. Compared to other methods, ours produces more accurate and stable results with details.}

\begin{figure*}
    \centering
    \includegraphics[width=\textwidth]{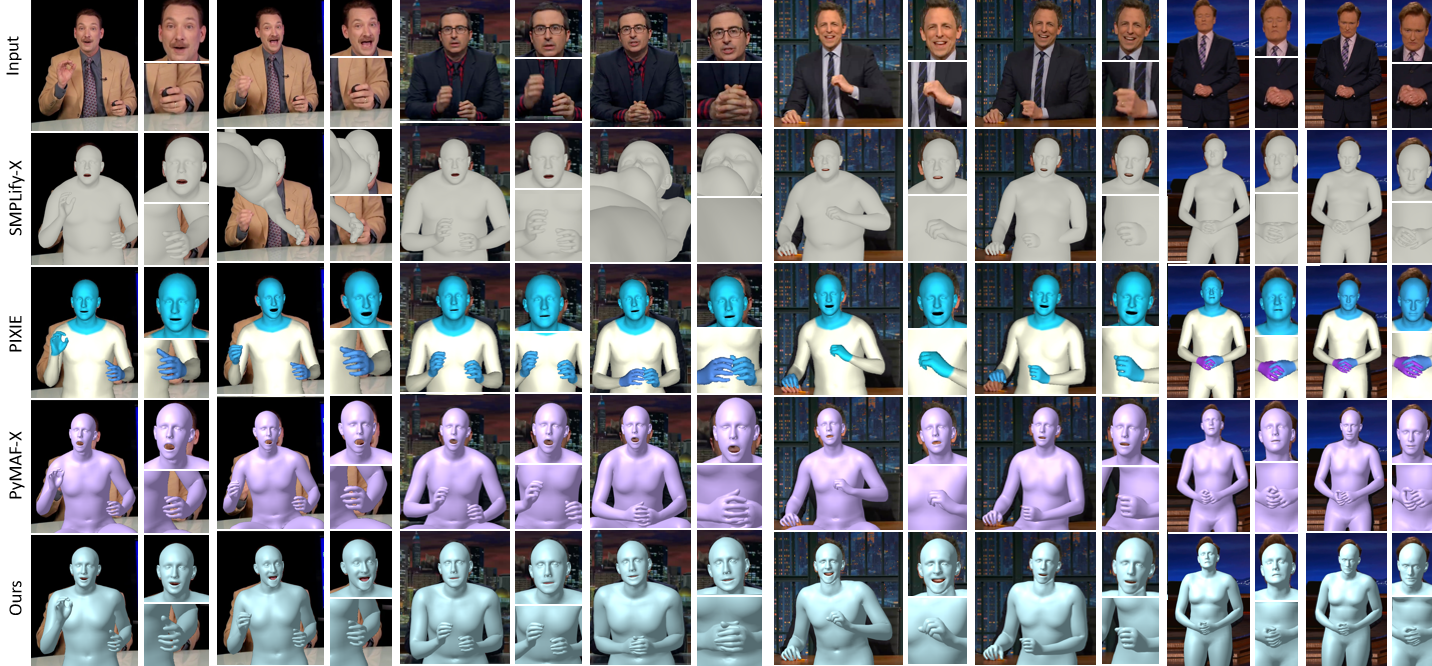}
    \caption{\reconstructedResultsCaptionv}
    \label{fig:rec_results}
\end{figure*}


\section{Network Architecture Details}

\subsection{Face Generator}
The raw audio input is normalized to zero mean and unit variance, and then is fed to encoder, which consists of an audio feature extractor, a transformer encoder, and a full-connected layer. The audio feature extractor is followed by an interpolation operation, in which the audio feature is re-sampled into target frames. For the decoder, it comprises six temporal convolution layers (with a kernel size, stride and padding of 3, 1 and 1 respectively) and a full-connected layer. Each temporal convolution layer is followed by layer normalization \cite{ba2016layer} and a Leaky RELU activation function \cite{maas2013rectifier}. We adopt SGD with momentum and a learning rate of 0.001 as the optimizer. The face generator is trained with batchsize of 1 for 100 epochs, in which each batch contains a full-length audio and corresponding facial motions.

\subsection{Body and Hand Generator}

\paragraph{VQ-VAEs Details.}
The VQ-VAE takes body or hand motions as input. The encoder of each VQ-VAE is composed of three residual layers, which includes three temporal convolution layers (with a kernel size, stride and padding of 3, 1 and 1 respectively) followed by batch normalization \cite{ioffe2015batch} and a Leaky RELU activation function \cite{maas2013rectifier}. The encoder is interleaved with a temporal convolution layer with a kernel size, stride and padding of 4, 2 and 1 respectively after every residual layer except the last so that the temporal window size $w$ is equal to 4. On the top of the encoder, a full-connected layer is added to reduce the dimension before quantization. The decoder is symmetric with the encoder. We adopt Adam with $\beta_1=0.9$, $\beta_2=0.999$ and a learning rate of 0.0001 as the optimizer. The commitment loss weight $\beta$ is set to 0.25. The VQ-VAEs are trained with a batchsize of 128 and a sequence length of 88 frames for 100 epochs. 

\paragraph{Autoregressive Model Details.}
The autoregressive model consists of an audio encoder and a Gated PixelCNN \cite{van2016conditional}. The audio encoder, which has the same structure as the VQ-VAE encoder, takes MFCC feature as input. Then we concatenate the output of the audio encoder and VQ-VAEs encoders and feed it to the Gated PixelCNN. The Gated PixelCNN has 15 gated convolution layers conditioned on identity, in which the convolution kernel is masked to make sure the model cannot read future information. We adopt Adam with $\beta_1=0.9$, $\beta_2=0.999$ and a learning rate of 0.0001 as the optimizer. The autoregressive model is trained with a batchsize of 128 and a sequence length of 88 frames for 100 epochs. 

\section{More Comparison}

\begin{figure}[t]
    \centering
    \includegraphics[width=\columnwidth]{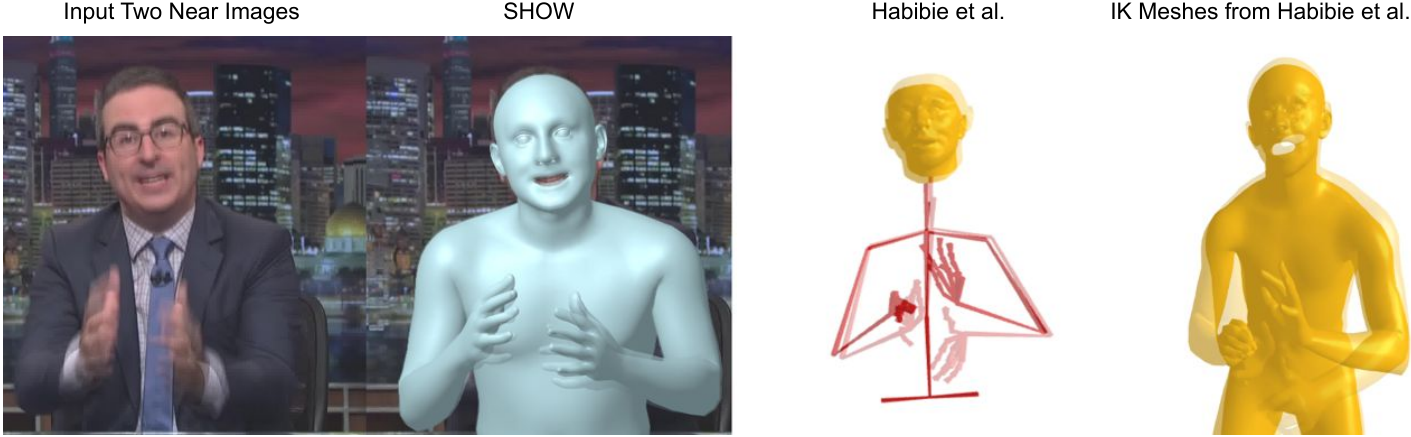}
    \vspace{-19pt}
    \caption{Holistic body reconstruction compared to Habibie et al.}
    \label{fig:rebuttal}
\end{figure}

\paragraph{Habibie el al. \citep{habibie2021learning} v.s. SHOW.}
Habibie et al. \citep{habibie2021learning} represent body, hands, and face separately.
The lack of connection between body and face/hands results in unnatural poses of the face/hands w.r.t.~the body. 
\Fref{fig:rebuttal} a) shows that the hand and head poses of the body mesh, reconstructed from their estimated 3D skeleton, are less accurate than ours. 
Generated video results are further jittery.
In contrast, SHOW generates more stable and accurate holistic body meshes.

\begin{table*}[t]
    \centering
    \footnotesize
    \begin{tabular}{l|ccccccc}
    \toprule
    \tabincell{l}{Method} 
    & \tabincell{c}{Habibie}
    & \tabincell{c}{Audio VAE}
    & \tabincell{c}{Audio+Motion VAE}
    & \tabincell{c}{Audio2Gesture}\citep{li2021audio2gestures}
    & \tabincell{c}{Ours w/o c-c}
    & \tabincell{c}{Ours w/ c-c}\\\hline
    FGD $\downarrow$     
    & 239.32
    & 121.01
    & 166.65
    & 203.99
    & 147.81
    & \textbf{74.88}\\ \hline
    Variance $\uparrow$     
    & 0
    & 0.044
    & 0.176
    & 0.240
    & \textbf{0.922}
    & 0.821\\ \hline
    BC (GT 0.868)     
    & 0.948
    & 0.746
    & 0.822
    & 0.943
    & 0.851
    & \textbf{0.872}\\ \hline
    \end{tabular}
    \caption{More experimental results.}
    \label{tab:new_metric}
\end{table*}

\paragraph{Experimental Results.}
We compare our method with more other approaches and more metrics in \tref{tab:new_metric}. 
Specifically, We add Frechet Gesture Distance (FGD) \citep{yoon2020speech} to measure the motion realism and beat consistency (BC) \citep{liu2022learning} to measure the alignment between the generated body motion and input audio, and compare with another audio-to-body motion baseline \citep{li2021audio2gestures}. 
Our method outperforms the baselines in all these metrics and generates more diverse body motions, which are better aligned with the input audio.

\section{Application}
One application of our speech-to-motion generation is to create the photo-realistic neural avatars through neural renderers such as SMPLpix~\cite{prokudin_smplpix_2020}. 
Given the mesh vertices provided by \speechmodelname and their colors, we first project them onto the image plane. Then, with the projected mesh vertices, SMPLpix allows us to efficiently synthesise photo-realistic images of humans. As \speechmodelname can produce continuous yet diverse motions, integrating SMPLpix with our motion generation framework enables us generate human avatars under different poses (see \fref{fig:application}), leading to end-to-end photo-realistic video generation.

\newcommand{\applicationCaption}{
The application with SMPLpix to create photo-realistic neural avatars. Top row (input): the mesh vertices provided by \speechmodelname and their colors projected onto the image plane, bottom row: rendered output.
}
\begin{figure}
    \centering
    \includegraphics[width=\columnwidth]{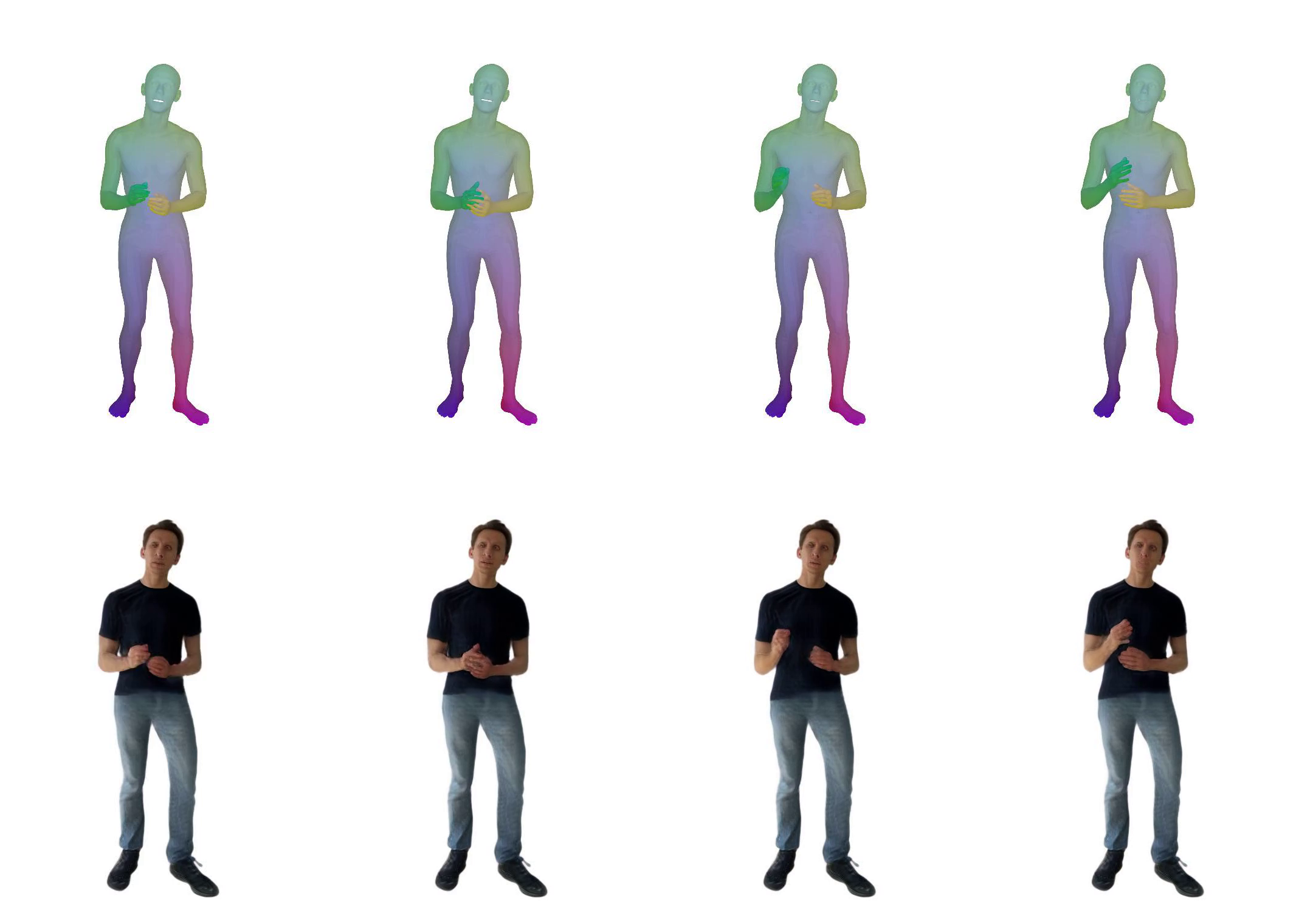}
    \caption{\applicationCaption}
    \label{fig:application}
\end{figure}


\section{Discussions}
\paragraph{Reconstruction.} \bodymethodname is based on SMPLify-X whose supervision signal is obtained from 2D keypoint reprojection. Thus, it is sensitive to severe hand shape deformation and heavy occlusion. A future direction would be to leverage advanced hand model with rich shape and pose space. Besides, \bodymethodname can only handle static camera cases currently. In the future, we plan to extend it to moving cameras. 



\paragraph{Audio2motion.} While we have demonstrated that \speechmodelname can generate realistic, coherent, and diverse holistic body motion with facial expression, body, and hand motions, it is subject to a limitation that can be addressed in the future. For the face generator, we mainly focus on facial motion (e.g. lip motion) and might not handle the very complex facial movements caused by emotions. In the future, we plan to extend to model this sort of part. 


\section{Risks and Potential Misuse}
This work is intended for studying the translation from human speech to holistic body motion, helping building virtual agents to behave realistically and interact with listeners meaningfully. Since our techniques can generate a realistic and diverse 3D talking humans from audio, there is a risk that such technique could be potentially misused for fake video generation. For instance, a fake speech could be used to construct highly realistic 3D holistic body motion while it never happened. Thus, we should use such technology responsibly and carefully. We hope to raise the public’s awareness about a safe use of such technology. 

\end{appendices}

\end{document}